\documentclass[10pt,twocolumn,letterpaper]{article}

\usepackage{iccv,times,graphicx,tabulary,multirow,subfig,xspace,xcolor}
\usepackage{amsthm,amssymb,fixmath,mathtools}
\usepackage[utf8]{inputenc}
\usepackage{amsmath}
\usepackage{amsfonts}
\usepackage{amssymb}
\usepackage{multirow}
\usepackage{color}
\usepackage{bm}
\usepackage{placeins}
\usepackage{pifont}
\usepackage{graphicx}
\usepackage{enumitem}
\usepackage{physics}

\usepackage{hyperref}
\hypersetup{
	colorlinks=true,
	linkcolor=blue,
	filecolor=magenta,      
	urlcolor=cyan,
	breaklinks=true,
	letterpaper=true,
	pagebackref=true,
	bookmarks=false
}

\graphicspath{{./figures/}}

\definecolor{citecolor}{RGB}{34,139,34}

\begin{document}
\title{Distribution-Aware Coordinate Representation of Keypoint for Human Pose Estimation}
\track{COCO 2019 Keypoint Detection}

\author{Hanbin Dai$^1$\thanks{equal contribution} 
\quad \quad \quad \quad Liangbo Zhou$^1$\footnotemark[1]
\quad \quad \quad \quad Feng Zhang$^1$\footnotemark[1]
\\
\quad \quad Zhengyu Zhang$^2$\footnotemark[1]
\quad \quad \quad \quad Hong Hu$^1$\footnotemark[1]
\quad \quad \quad \quad Xiatian Zhu$^3$\footnotemark[1]
\quad \quad \quad \quad Mao Ye$^1$
\\
University of Electronic Science and Technology of China$^1$ \\
\quad 
Shenzhen University$^2$
\quad 
University of Surrey$^3$\\
{\tt\small \{daihanbin.ac, heathhose96, zhangfengwcy\}@gmail.com} \\
\tt\small zhengyuzhang23@outlook.com
\quad
\tt\small Andrewhuhong@gmail.com
\quad
\tt\small eddy.zhuxt@gmail.com
\quad
\tt\small cvlab.uestc@gmail.com
}

\maketitle

\begin{abstract}
	In this paper, we focus on the coordinate representation in human pose estimation. 
	While being the standard choice, heatmap based representation has not been systematically
	investigated. We found that the process of coordinate decoding
	(i.e. transforming the predicted heatmaps to the coordinates)
	is surprisingly significant for human pose estimation performance, which nevertheless was not recognised before. In light of the discovered importance, we further probe the design limitations of the standard coordinate decoding method and propose a principled distribution-aware decoding method. Meanwhile, we
	improve the standard coordinate encoding process (i.e. transforming ground-truth coordinates to heatmaps) by generating
	accurate heatmap distributions for unbiased model training.
	Taking them together, we formulate a novel Distribution-Aware coordinate Representation for Keypoint (DARK) method. Serving as a model-agnostic plug-in, DARK significantly improves
	the performance of a variety of state-of-the-art human pose
	estimation models. Extensive experiments show that DARK
	yields the best results on COCO keypoint detection challenge, validating the usefulness and effectiveness of our novel coordinate representation idea. The project page containing more details is at \url{https://ilovepose.github.io/coco/}
\end{abstract}

\section{Introduction}
Human pose estimation is a challenging problem in computer vision aiming at finding the coordinates of human body parts.
Recently, convolutional neural networks (CNNs) have achieved significant success \cite{toshev2014deeppose,pfister2015flowing,wei2016convolutional,newell2016stacked,yang2017pyramid,xiao2018simple,sun2019deep}.
However, these methods 
typically focus on designing pose specific architecture, ignoring the coordinate representation of body parts.
In the classification task, the one-hot vectors are utilised to represent the object class, so that the model can learn the target easily.
A human pose estimation model also needs a target representation (coordinate encoding and decoding).
The {\em de facto} standard coordinate representation of body part is coordinate heatmap generated using a 2D Gaussian distribution/kernel centred at the labelled coordinate of each joint \cite{tompson2014joint}.
Down-sampling is often needed for controlling 
the computational cost.

In the literature, the problem of coordinate encoding and
decoding (i.e. denoted as coordinate representation) gains
little attention, although being indispensable in model training and inference. 
Contrary to the existing human pose estimation studies,
in this work we dedicatedly investigate the problem of joint
coordinate representation including encoding and decoding.
Moreover, we recognise that the heatmap resolution is one
major obstacle that prevents the use of smaller input resolution for faster model inference. 
In light of the discovered significance of coordinate representation, we conduct in-depth investigation and recognise
that one key limitation lies in the coordinate decoding process. Whilst existing standard shifting operation has shown
to be effective as found in this study, we propose a principled
distribution-aware representation method for more accurate
joint localisation at sub-pixel accuracy. Specifically, it is designed to comprehensively account for the distribution information of heatmap activation via Taylor-expansion based
distribution approximation. Besides, we observe that the
standard method for generating the ground-truth heatmaps
suffers from quantisation/discretisation errors, leading to imprecise supervision and inferior performance. To solve
this issue, we propose generating unbiased heatmaps allowing Gaussian kernel being centred at sub-pixel locations.

The contribution of this work is that, we discover the previously unrealised significance of coordinate representation
in human pose estimation, and propose a novel Distribution-Aware coordinate Representation for Keypoint (DARK) method with two
key components: (1) efficient Taylor-expansion based coordinate decoding, and (2) unbiased sub-pixel centred coordinate encoding. Importantly, existing human pose methods
can be seamlessly benefited from DARK without any algorithmic modification. 
Extensive experiments on COCO keypoint benchmark show that our method
provides significant performance gain for the existing
state-of-the-art human pose estimation model \cite{newell2016stacked, xiao2018simple, sun2019deep}, achieving the best single model accuracy. DARK favourably enables the use of smaller input
image resolutions with much smaller performance degradation, whilst dramatically boosting the model inference efficiency. %

\begin{figure*}%
	\centering
	\includegraphics[width=0.95\linewidth]{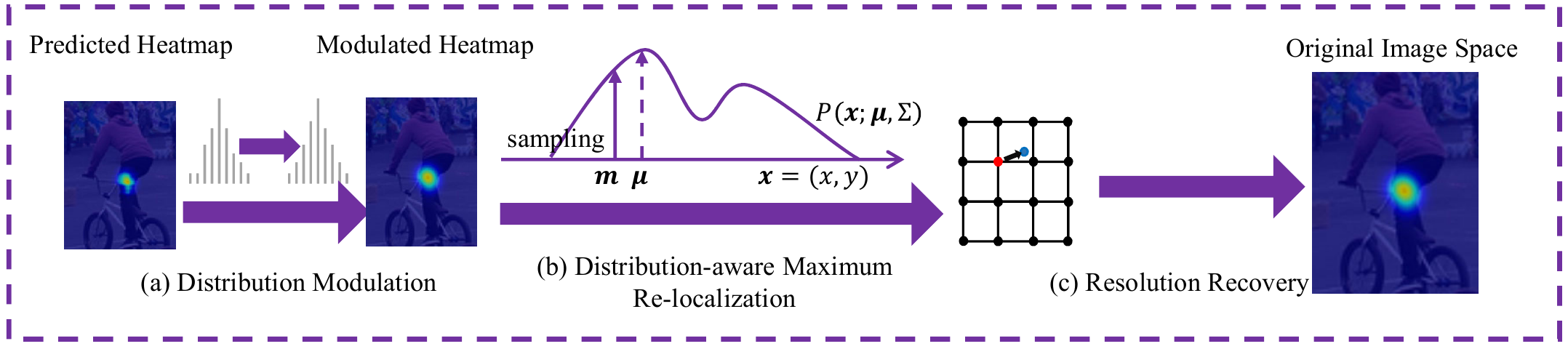}
	\caption{Overview of the proposed distribution aware coordinate decoding method.
	}
	\label{fig:test_stage}
\end{figure*}

\section{Method}

\subsection{Human Pose Estimator}
We find a significant performance bottleneck in the coordinate representation (coordinate encoding and decoding), and 
introduce a principled solution, named as
{\em Distribution-Aware coordinate Representation for Keypoint}
(DARK).
In the following we first describe the decoding process,
focusing on the limitation analysis of the existing standard
method and the development of a novel solution. Then, we discuss and address the limitations of the encoding
process.

\subsubsection{Coordinate Decoding}
Suppose a pose estimator outputs a heatmap matching the spatial size of an input image.
It is easy to obtain the location of the body joints by identifying the maximum activation in the heatmap.
However, this is often not the case due to the computation budget constraint. 
Instead, we need to upscale the low-resolution heatmap to the original image resolution.
This involves a sub-pixel localisation problem.
The standard method is to offset the max-activation prediction by a quarter of a pixel in the direction towards the second
max activation before transforming back to the original coordinate space of the input image.
This hand-designed method is not sufficiently accurate without
good insights.

To solve the sub-pixel localisation problem, 
we propose a Taylor-expansion based re-localisation method, 
called distribution-aware
maximum re-localisation (Fig \ref{fig:test_stage} (b)).
Specifically, we exploit the Taylor-expansion theory
to estimate the underlying max activation
in a Gaussian distribution assumption.
The predicted Gaussian heatmap is often ill-conditioned
which may hurt the offset estimation.
We therefore further design a heatmap distribution modulation method (Fig \ref{fig:test_stage}(a)) for preprocessing.
Specifically, a Gaussian kernel is utilised to smooth the predicted heatmap. %

\subsubsection{Coordinate Encoding}
The heatmap based representation assume the coordinate of a body part follows a 2D Gaussian distribution.
In the coordinate encoding phase, the original person images is downsampled into the model
input size. So, the ground-truth joint coordinates require to
be transformed accordingly before generating the heatmaps.

Formally, we denote by $\bm{g}=(u,v)$ the ground-truth coordinate of a joint.
The resolution reduction is defined as:
\begin{align}
\bm{g}' =(u',v') = \frac{\bm{g}}{\lambda} = (\frac{{u}}{\lambda}, \frac{{v}}{\lambda})
\end{align}
where $\lambda$ is the downsampling ratio.

Conventionally, for facilitating the kernel generation,
we often quantise $\bm{g}'$:
\begin{align}
\bm{g}'' =(u'',v'') = \text{quantise}(\bm{g}') = \text{quantise}(\frac{{u}}{\lambda}, \frac{{v}}{\lambda})
\label{eq:quantise}
\end{align}
where $\text{quantise}()$ specifies a quantisation function, with the common choices including
floor, ceil and round.  

Subsequently, the heatmap centred at the quantised coordinate $\bm{g}''$ can be
synthesised through:
\begin{equation}\small
\mathcal{G} (x, y;\bm{g}'' ) = 
\frac{1}{ 2\pi \sigma^2 }
\exp( -\frac{(x-u'')^{2} + (y-v'')^2}{2\sigma^2}  )
\label{eq:heatmap_gt}
\end{equation}
where $(x,y)$ specifies a pixel location in the heatmap,
and $\sigma$ denotes a fixed spatial variance.

Obviously, the heatmaps generated in the above way
are {\em inaccurate} and {\em biased} due to the quantisation error.
This may introduce sub-optimal supervision signals
and result in degraded model performance,
particularly for the case of accurate coordinate encoding
as proposed in this work.

To address this issue, we simply place the heatmap centre
at the non-quantised location $\bm{g}'$ which represents
the {\em accurate} ground-truth coordinate.
We still apply Eq. \eqref{eq:heatmap_gt} 
but replacing $\bm{g}''$ with $\bm{g}'$.

\subsection{Person Detection}
To obtain good person detection results efficiently, 
we use the Hybrid Task Cascade (HTC) detector \cite{chen2019hybrid} and the SNIPER detector \cite{sniper2018} jointly
for the challenge entry model.

\section{Experiments}
\subsection{Datasets}
We used two datasets.
(1) The {\bf COCO} keypoint dataset \cite{lin2014microsoft} consists of about 200K images containing 250K person instances labelled with 17 joints.
It has four splits: train, val, test-dev, test-challenge with 118K, 5K, 20K and 20K images respectively. 
(2) The {\bf AIC} dataset \cite{aic2017} contains about 300k images and 700k person instance labelled with 14 keypoints.
It has four splits: train, val, test A, test B with 
210K, 30K, 30K and 30K images respectively.

\subsection{Ablation Study}
\subsubsection{Evaluating Coordinate Representation}
In this test,
we used the person detection results from \cite{sun2019deep}.
By default we used HRNet-W32 as the backbone model and 128x96 as the input size, and reported the accuracy results on the COCO validation set.

\vspace{0.1cm}
\noindent{\bf (i) Coordinate decoding }
We evaluated the proposed coordinate decoding.
The conventional biased heatmaps were used.
We compared the proposed distribution-aware shifting method with
{\em no shifting} (\ie directly using the maximal activation location),
and the {\em standard shifting} in \cite{newell2016stacked, chen2018cascaded, xiao2018simple, sun2019deep}.
We observed in Table \ref{tbl:coord_decod} that:
{\bf (i)} The standard shifting
gives as high as 5.7\% AP accuracy boost,
which is surprisingly effective.
This reveals previously unseen significance of coordinate
decoding to human pose estimation.
{\bf (ii)} Despite the great gain by the standard decoding method,
the proposed model further improves AP score by 1.5\%.

\vspace{0.1cm}
\noindent{\bf (ii) Coordinate encoding }
We compared the proposed {\em unbiased} encoding 
with the standard {\em biased} encoding,
along with both the standard and our decoding method.
We observed from Table \ref{tbl:coord_encode}
that our unbiased encoding with accurate kernel centre
brings positive performance margin, regardless of 
the coordinate decoding method.

\begin{table}%
	\setlength{\tabcolsep}{0.1cm}
	\begin{center}
		\caption{
			Effect of coordinate decoding on COCO val.
			Model: HRNet-W32;
			Input size: $128\times96$.
		}
		\label{tbl:coord_decod}
		\begin{tabular}{ c | c |c | c | c | c | c  }
			\hline
			Decoding & $AP$ & $AP^{50}$ & $AP^{75}$ & $AP^{M}$ & $AP^{L}$ & $AR$ \\
			\hline \hline
			No Shifting
			& 61.2 & 88.1 & 72.3 & 59.0 & 66.3 & 68.7 
			\\
			Standard Shifting
			& 66.9 &\bf 88.7 & 76.3 & 64.6 & 72.3 & 73.7 
			\\ 
			\hline
			\bf Ours
			& \bf 68.4 & 88.6 & \bf 77.4 & \bf 66.0 & \bf 74.0 & \bf 74.9
			\\
			\hline
		\end{tabular}
	\end{center}

\end{table}

\begin{table} %
	\setlength{\tabcolsep}{0.1cm}
	\begin{center}
		\caption{
			Effect of coordinate encoding on COCO val.
			Model: HRNet-W32;
			Input size: $128\times96$.
		}
		\label{tbl:coord_encode}
		\resizebox{\columnwidth}{!}{%
			\begin{tabular}{ c| c | c |c | c | c | c | c  }
				\hline
				Encode & Decode & $AP$ & $AP^{50}$ & $AP^{75}$ & $AP^{M}$ & $AP^{L}$ & $AR$ \\
				\hline \hline
				Biased & Standard
				& 66.9 & 88.7 & 76.3 & 64.6 & 72.3 & 73.7 
				\\
				\bf Unbiased & Standard
				& \bf 68.0 & \bf 88.9  & \bf 77.0  & \bf 65.4  & \bf 73.7  & \bf 74.5
				\\
				\hline \hline
				Biased &\bf Ours
				& 68.4 & 88.6 & 77.4 & 66.0 & 74.0 & 74.9
				\\
				\bf Unbiased &\bf Ours
				& \bf 70.7 & \bf  88.9 & \bf  78.4 & \bf  67.9 & \bf  76.6 & \bf  76.7
				\\
				\hline
			\end{tabular}
		}
	\end{center}
\end{table}

\vspace{0.1cm}
\noindent{\bf (iii) Input resolution }
We examined the impact of input image resolution/size.
We compared our DARK model 
(HRNet-W32 as backbone) 
with the original HRNet-W32 using the 
biased heatmap supervision for training and 
the standard shifting for testing.
From Table \ref{tbl:ablation_input_size} we have a couple of observations:
{\bf (a)} With reduced input image size, as expected the model performance consistently 
degrades whilst the inference cost drops clearly.
{\bf (b)} With the support of DARK,
the model performance loss can be effectively mitigated,
especially in case of very small input resolution (\ie very fast model inference).
\begin{table} %
	\setlength{\tabcolsep}{0.1cm}
	\begin{center}
		\caption{
			Effect of input image size on COCO val.
			DARK uses HRNet-W32 (HRN32) as backbone.
		}
		\label{tbl:ablation_input_size}
		\resizebox{\columnwidth}{!}{%
			\begin{tabular}{ c | c | c | c | c | c | c | c | c  }
				\hline
				Method & Input size & GFLOPs & $AP$ & $AP^{50}$ & $AP^{75}$ & $AP^{M}$ & $AP^{L}$ & $AR$ \\
				\hline \hline
				HRN32 & \multirow{2}{*}{128$\times$96} & \multirow{2}{*}{1.8}
				& 66.9 & 88.7 & 76.3 & 64.6 & 72.3 & 73.7 
				\\
				\bf DARK &  &
				& \bf 70.7 & \bf 88.9 & \bf 78.4 & \bf 67.9 & \bf 76.6 & \bf 76.7
				\\ \hline
				HRN32 & \multirow{2}{*}{256$\times$192} & \multirow{2}{*}{7.1}
				& 74.4	& \bf 90.5	& 81.9	& 70.8	& 81.0	& 79.8
				\\ 
				\bf DARK & &
				& \bf 75.6 & \bf 90.5 & \bf 82.1 & \bf 71.8 & \bf 82.8 & \bf 80.8
				\\ \hline
				HRN32 & \multirow{2}{*}{384$\times$288} & \multirow{2}{*}{16.0}
				& 75.8	& 90.6 & 82.5 & 72.0 & 82.7 & 80.9
				\\ 
				\bf DARK & & 
				& \bf 76.6 & \bf 90.7 & \bf 82.8 & \bf 72.7 & \bf 83.9 & \bf 81.5
				\\ 
				\hline
			\end{tabular}
		}
	\end{center}
\end{table}

\subsubsection{Effect of DARK}
We further evaluate the effect of DARK on the COCO test-dev set.
We compared the HRNet-W48(HRN48) with DARK using HRNet-W48 backbone.
We observed from Table \ref{tbl:dark} that DARK 
gives a clear performance gain.
\begin{table}%
	\setlength{\tabcolsep}{0.1cm}
	\begin{center}
		\caption{
			Effect of DARK on COCO test-dev;
			Input size: $384\times288$;
			Training data: COCO train;
			Detection: MSRA\cite{sun2019deep}.
		}
		\label{tbl:dark}
		\begin{tabular}{ c | c |c | c | c | c | c  }
			\hline
			Method & $AP$ & $AP^{50}$ & $AP^{75}$ & $AP^{M}$ & $AP^{L}$ & $AR$ \\
			\hline \hline
			HRN48 &
			75.5 & \bf 92.5 & 83.3 & 71.9 & 81.5 & 80.5 
			\\
			DARK &
			\bf 76.2 & \bf 92.5 & \bf 83.6 & \bf 72.5 & \bf 82.4 & \bf 81.1
			\\
			\hline
		\end{tabular}
	\end{center}
\end{table}

\subsubsection{Effect of Extra Training Data}
We examined the impact of extra training data
with DARK(HRNet-W48) from AIC.
The results in Table \ref{tbl:dataset} show that extra training data brings a positive performance boost, as expected.

\begin{table}%
	\setlength{\tabcolsep}{0.1cm}
	\begin{center}
		\caption{
			Effect of extra training data on COCO test-dev;
			Model: DARK using HRNet-W48 as backbone;
			Input size: $384\times288$;
			Detection: MSRA\cite{sun2019deep}.
		}
		\label{tbl:dataset}
		\begin{tabular}{ c | c |c | c | c | c | c  }
			\hline
			Dataset & $AP$ & $AP^{50}$ & $AP^{75}$ & $AP^{M}$ & $AP^{L}$ & $AR$ \\
			\hline \hline
			COCO &
			76.2 & 92.5 & 83.6 & 72.5 & 82.4 & 81.1
			\\
			COCO+AIC &
			\bf 77.4 & \bf 92.6 & \bf 84.6 & \bf 73.6 & \bf 83.7 & \bf 82.3
			\\
			\hline
		\end{tabular}
	\end{center}
\end{table}

\subsubsection{Effect of Person Detection}
We examined different person detectors. %
We observed in Table \ref{tbl:detector} that:
{\bf (i)}  HTC is the best detector;
{\bf (ii)} the combined detection can boost 
the overall performance. %

\begin{table}%
	\setlength{\tabcolsep}{0.1cm}
	\begin{center}
		\caption{
			Effect of person detection on COCO test-dev;
			Model: DARK using HRNet-W48 as backbone;
			Input size: $384\times288$;
			Training data: COCO train+AIC train;
			Detection: MSRA\cite{sun2019deep}, HTC\cite{chen2019hybrid}, SNIPER\cite{sniper2018}.
		}
		\label{tbl:detector}
		\begin{tabular}{ c | c |c | c | c | c | c  }
			\hline
			Detector & $AP$ & $AP^{50}$ & $AP^{75}$ & $AP^{M}$ & $AP^{L}$ & $AR$ \\
			\hline \hline
			MSRA &
			77.4 & 92.6 & 84.6 & 73.6 & 83.7 & 82.3
			\\
			\hline
			SNIPER &
			78.0 & \bf 93.6 & 85.1 & 74.2 & 83.6 & 82.6
			\\
			HTC &
			\bf78.2 & 93.5 & \bf 85.5 &\bf 74.4 & 83.7 & 83.2
			\\
			\hline
			HTC+SNIPER & \bf 78.2 & 93.5 & \bf 85.5 & \bf 74.4 & \bf 84.2 & \bf 83.5
			\\
			\hline
		\end{tabular}
	\end{center}
\end{table}

\subsubsection{Effect of Model Ensemble}
We formed two ensembles:
one with 3 models and one with 8 models.
They were trained by DARK
with varying backbones (HRNet-W48, HRNet-W32, ResNet-152),
training data (COCO, COCO+AIC),
and batch sizes (small and large).
Table \ref{tbl:single}
shows that model ensemble helps.

\begin{table}%
	\setlength{\tabcolsep}{0.1cm}
	\begin{center}
		\caption{
		    Best single model vs. ensemble of 3/8 models
			on COCO test-dev;
			Input size: $384\times288$;
			Detection: HTC+SNIPER;
		}
		\label{tbl:single}
		\begin{tabular}{ c | c |c | c | c | c | c  }
			\hline
			Model & $AP$ & $AP^{50}$ & $AP^{75}$ & $AP^{M}$ & $AP^{L}$ & $AR$ \\
			\hline
			Best Single &
			78.2 & 93.5 & 85.5 & 74.4 & 84.2 & 83.5
			\\
			Ensemble(3) &
			78.7 & 93.6 & \bf 86.0 & 74.7 & 84.3 & \bf 83.5
			\\
			Ensemble(8) & \bf 78.9 & \bf 93.8 & \bf 86.0 & \bf 75.1 & \bf 84.4 & \bf 83.5
			\\
			\hline
		\end{tabular}
	\end{center}
\end{table}

\subsection{ICCV Keypoint Detection Challenge }
We used an ensemble of 8 DARK models for the challenge.
Table \ref{tbl:challenge} shows that 
our method achieves 76.4\% AP
for multi-person pose estimation on COCO test-challenge set.
\begin{table}%
	\setlength{\tabcolsep}{0.1cm}
	\begin{center}
		\caption{
			Result of 8-model ensemble on COCO test-challenge.
		}
		\label{tbl:challenge}
		\begin{tabular}{ c |c | c | c | c | c  }
			\hline
			 $AP$ & $AP^{50}$ & $AP^{75}$ & $AP^{M}$ & $AP^{L}$ & $AR$ \\
			\hline
			76.4 & 92.5 & 82.7 & 70.9 & 83.8 & 81.6
			\\
			\hline
		\end{tabular}
	\end{center}
\end{table}

\section{Conclusion}
We presented a strong human pose estimation method
based on a novel distribution-aware coordinate representation idea. It achieves very competitive results on COCO keypoint detection challenge.
Please visit our project page for more details.

{\small \bibliographystyle{ieee_fullname} \bibliography{references}}

\end{document}